\documentclass[letterpaper]{article} 
\usepackage{aaai23}  
\usepackage{times}  
\usepackage{helvet}  
\usepackage{courier}  
\usepackage[hyphens]{url}  
\usepackage{graphicx} 
\usepackage{natbib}  
\usepackage{caption} 
\usepackage{algorithm}
\usepackage{algorithmic}
\usepackage[inkscapelatex=false]{svg}

\usepackage{booktabs}
\usepackage{multirow}
\usepackage{amsmath}
\usepackage{amssymb}
\usepackage{subcaption}
\usepackage{color}
\usepackage{enumitem}

\usepackage{newfloat}
\usepackage{listings}
\DeclareCaptionStyle{ruled}{labelfont=normalfont,labelsep=colon,strut=off} 
\floatstyle{ruled}
\newfloat{listing}{tb}{lst}{}
\floatname{listing}{Listing}
\title{Universal Information Extraction as Unified Semantic Matching}
\author {
    Jie Lou\textsuperscript{\rm 1}\thanks{~ Equally contribution.},
    Yaojie Lu\textsuperscript{\rm 2}\footnotemark[1],
    Dai Dai\textsuperscript{\rm 1}\thanks{~ Corresponding authors.},
    Wei Jia\textsuperscript{\rm 1},
    Hongyu Lin\textsuperscript{\rm 2}, \\
    Xianpei Han\textsuperscript{\rm 2,3}\footnotemark[2],
    Le Sun\textsuperscript{\rm 2,3},
    Hua Wu\textsuperscript{\rm 1}
}
\affiliations {
    \textsuperscript{\rm 1}Baidu Inc., Beijing, China\\
    \textsuperscript{\rm 2}Chinese Information Processing Laboratory ~ \textsuperscript{\rm 3}State Key Laboratory of Computer Science \\
Institute of Software, Chinese Academy of Sciences, Beijing, China\\
    \{loujie, daidai, jiawei07, wu\_hua\}@baidu.com \\
    \{luyaojie, hongyu, xianpei, sunle\}@iscas.ac.cn
}

\usepackage{bibentry}

\begin{document}

\maketitle

\begin{abstract}

The challenge of information extraction (IE) lies in the diversity of label schemas and the heterogeneity of structures.
Traditional methods require task-specific model design and rely heavily on expensive supervision, making them difficult to generalize to new schemas.
In this paper, we decouple IE into two basic abilities, structuring and conceptualizing, which are shared by different tasks and schemas.
Based on this paradigm, we propose to universally model various IE tasks with \textbf{U}nified \textbf{S}emantic \textbf{M}atching (USM) framework, which introduces three unified token linking operations to model the abilities of structuring and conceptualizing.
In this way, USM can jointly encode schema and input text, uniformly extract substructures in parallel, and controllably decode target structures on demand.
Empirical evaluation on 4 IE tasks shows that the proposed method achieves state-of-the-art performance under the supervised experiments and shows strong generalization ability in zero/few-shot transfer settings.

\end{abstract}

\section{Introduction} \label{sec:intro}

Information extraction aims to extract various information structures from texts \citep{andersen-etal-1992-automatic,grishman_2019}.
For example, given the sentence ``Monet was born in Paris, the capital of France'', an IE system needs to extract various task structures such as entities, relations, events, or sentiments in the sentence.
It is challenging because the target structures have diversified label schemas (person, work for, positive sentiment, etc.) and heterogeneous structures (span, triplet, etc.).

Traditional IE model leverages task- and schema-specialized architecture, which is commonly specific to different target structures and label schemas.
The expensive annotation leads to limited predefined categories and small data size in general domains for information extraction tasks.
From another perspective, task-specific model design makes it challenging to migrate learned knowledge between different tasks and extraction frameworks.
The above problems lead to the poor performance of IE models in low-resource settings or facing new label schema, which greatly restricts the application of IE in real scenarios.

Very recently, ~\citet{lu-etal-2022-unified} proposed the concept of universal information extraction (UIE), which aims to resolve multiple IE tasks using one universal model. To this end, they proposed a sequence-to-sequence generation model, which takes flattened schema and text as input, and directly generates diversified target information structures. Unfortunately, all associations between information pieces and schemas are implicitly formulated due to the black-box nature of sequence-to-sequence models \citep{alvarez-melis-jaakkola-2017-causal}. Consequently, it is difficult to identify what kind of abilities and knowledge are learned to transfer across different tasks and schemas. 
Therefore we have no way of diagnosing under what circumstances such transfer learning across tasks or schemas would fail.
For the above reasons, it is necessary to explicitly model and learn transferable knowledge to obtain effective, robust, and explainable transferability.

\begin{figure}[!tpb]
    \setlength{\belowcaptionskip}{-8pt}
    \includegraphics[width=0.48 \textwidth]{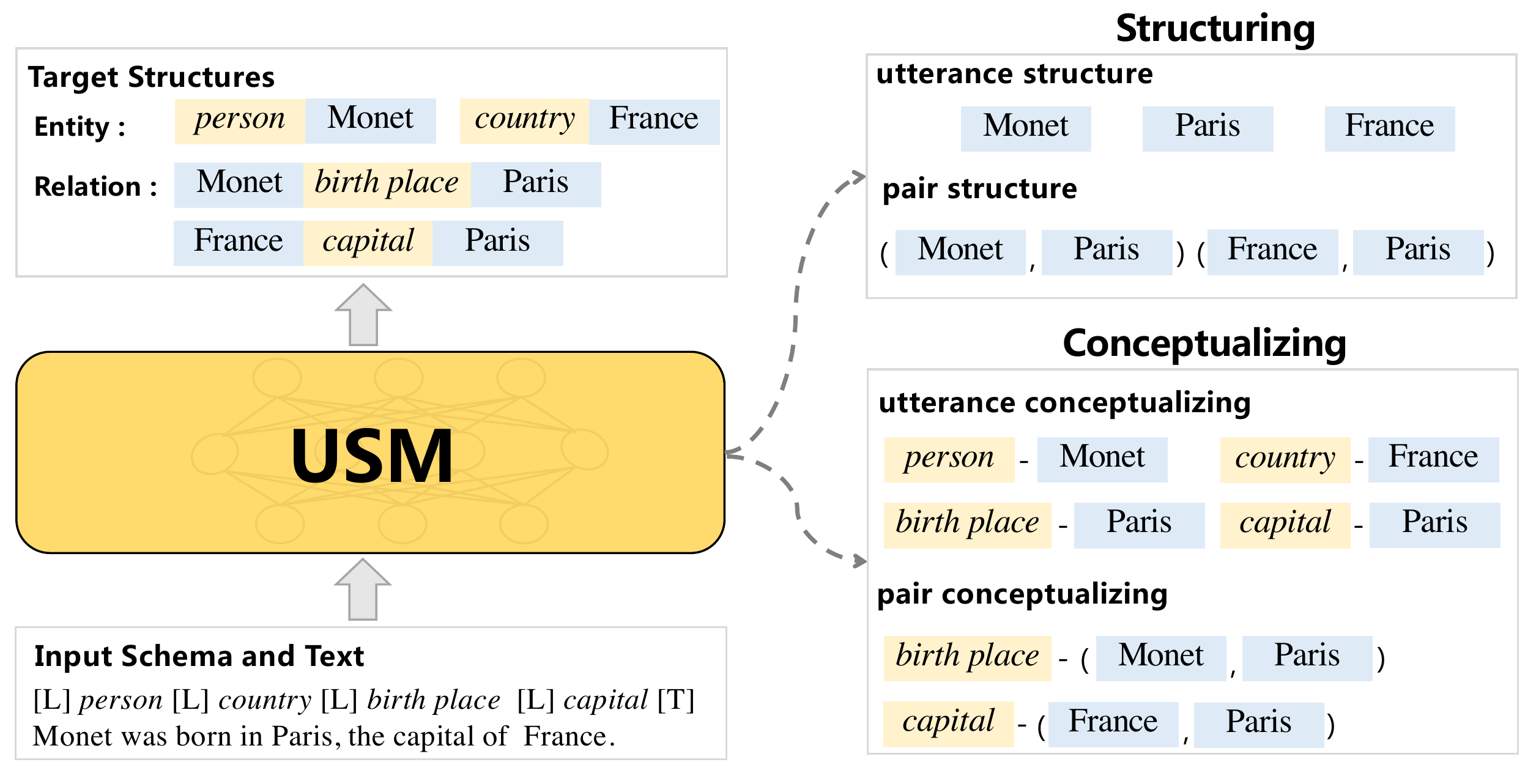}
    \caption{The USM framework for UIE.
    USM takes label schema and text as input and directly outputs the target structure through the \textbf{\textit{Structuring}} and \textbf{\textit{Conceptualizing}} operations.}
    \label{figure:motivation}
\end{figure}

\begin{figure*}[!tpb]
\centering
    \setlength{\belowcaptionskip}{-0.3cm}

\includegraphics[width=0.95\textwidth]{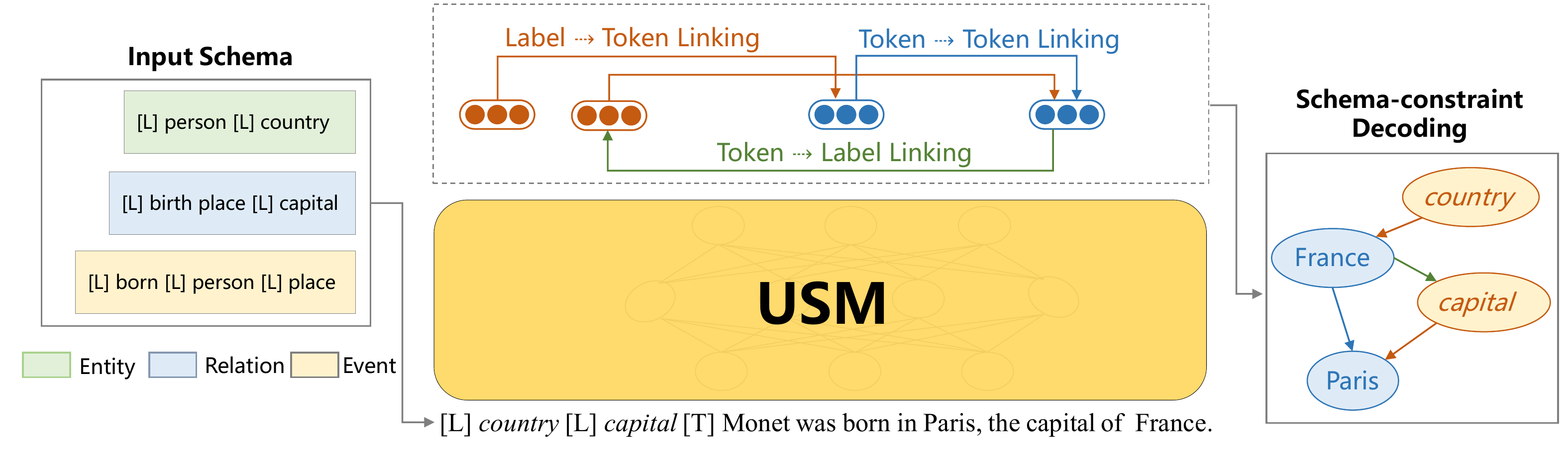}
\caption{The overall framework of Unified Semantic Matching.}
\label{figure:framework}
\end{figure*}

We find that, as shown in \figurename~\ref{figure:motivation}, even with diversified tasks and extraction targets, all IE tasks can be fundamentally decoupled into the following two critical operations:
1) \textit{\textbf{Structuring}}, which proposes label-agnostic basic substructures of the target structure from the text.
For example, proposing the utterance structure ``Monet'' for entity mention and ``born in'' for event mention, the associated pair structure (``Monet'', ``Paris'') for relation mention, and (``born in'', ``Paris'') for event argument mention.
2) \textit{\textbf{Conceptualizing}}, which generalizes utterance and paired substructures to corresponding target semantic concepts.
More importantly, these two operations can be explicitly reformulated using a semantic matching paradigm when given a target extraction schema. Specifically, structuring operations can be viewed as building specific kinds of semantic associations between utterances in the input text, while conceptualizing operations can be regarded as matching between target semantic labels and the given utterances or substructures. Consequently, if we universally transform information extraction into combinations of a series of structuring and conceptualizing, reformulate all these operations with the semantic matching between structures and schemas, and jointly learn all IE tasks under the same paradigm, we can easily conduct various kinds of IE tasks with one universal architecture and share knowledge across different tasks and schemas.

Unfortunately, directly conducting semantic matching between structures and schemas is impractical for universal information extraction.
First, sentences have many substructures, resulting in a large number of potential matching candidates and a large scale of matching, which makes the computational efficiency of the model unacceptable.
Second, the schema of IE is structural and hard to match with the plain text.
In this paper, we propose directed token linking for universal IE.
The main idea is to transform the structuring and conceptualizing into a series of directed token linking operations, which can be reverted to semantic matching between utterances and schema.

Based on the above observation, we propose USM, a unified semantic matching framework for universal information extraction (UIE), which decomposes structures and verbalizes label types for sharing structuring and conceptualizing abilities.
Specifically, we design a set of directed token linking operations (token-token linking, label-token linking, and token-label linking) to decouple task-specific IE tasks into two extraction abilities.
To learn the common extraction abilities, we pre-train USM by leveraging heterogeneous supervision from linguistic resources.
Compared to previous works, USM is a new transferable, controllable, efficient end-to-end framework for UIE, which jointly encodes extraction schema and input text, uniformly extracts substructures, and controllably decodes target structures on demand.

We conduct experiments on four main IE tasks under the supervised, multi-task, and zero/few-shot transfer settings.
The proposed USM framework achieves state-of-the-art results in all settings and solves massive tasks using a single multi-task model.
Under the zero/few-shot transfer settings, USM shows a strong cross-type transfer ability due to the shared structuring and conceptualizing obtained by pre-training.

In summary, the main contributions of this paper are:
\begin{enumerate} 
    \item
    We propose an end-to-end framework for universal information extraction – USM, which can jointly model schema and text, uniformly extract substructures, and controllably generate the target structure on demand.
    \item
    We design three unified token linking operations to decouple various IE tasks, sharing extraction capabilities across different target structures and semantic schemas and achieving ``one model for solving all tasks'' by multi-task learning.
    \item
    We pre-train a universal foundation model with large-scale heterogeneous supervisions, which can benefit future research on IE.
\end{enumerate}

\begin{figure*}[ht]
\centering
    \setlength{\belowcaptionskip}{-0.3cm}

\includegraphics[width=0.98 \textwidth]{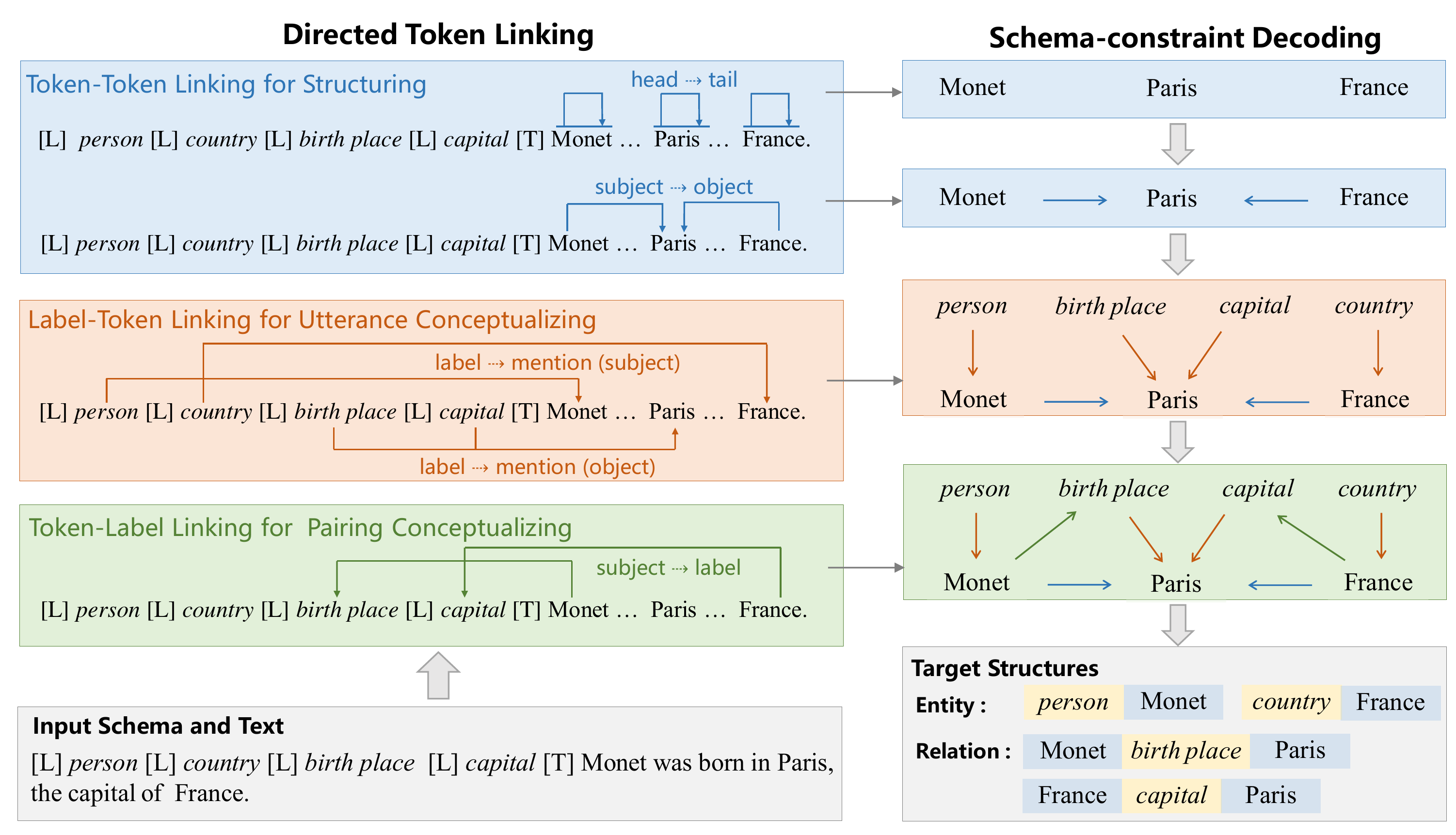}
\caption{
    Illustrations of Directed Token Linking.
    Token-Token Linking structures utterance and association pair substructures from the text, Label-Token Linking conceptualizes the utterance, and Token-Label Linking conceptualizes the association pair.
    In practice, we employ different label symbols ``\texttt{[L]}'' for utterance conceptualizing:
    ``\texttt{[LM]}'' for the label of single mention, such as entity types and event trigger types;
    ``\texttt{[LP]}'' for the predicate of association pair, such as relation types and event argument types.
}
\label{figure:unified-token-matching}
\end{figure*}

\section{Unified Semantic Matching via Directed Token Linking} \label{sec:model}

Information extraction is structuring the text's information and elevating it into specific semantic categories.
As shown in \figurename~\ref{figure:framework}, USM takes the arbitrary extraction label schema $l$ and the raw text $t$ as input and directly outputs the structure according to the given schema.
For example, given the text ``Monet was born in Paris,
the capital of France'', USM needs to extract (``France'', \textit{capital}, ``Paris'') for the relation type \textit{capital} and (\textit{person}, ``Monet'')/(\textit{country}, ``France'') for the entity type \textit{person} and \textit{country}.
The main challenges here are:
1) how to unifiedly extract heterogeneous structures using the shared structuring ability;
2) how to uniformly represent different extraction tasks under diversified label schemas to share the common conceptualizing ability.

In this section, we describe how to end-to-end extract the information structures from the text using USM.
Specifically, as shown in \figurename~\ref{figure:unified-token-matching}, USM first verbalizes all label schemas \citep{levy-etal-2017-zero,li-etal-2020-unified,lu-etal-2022-unified} and learns the schema-text joint embedding to build a shared label text semantic space.
Then we describe three basic token linking operations and how to structure and conceptualize information from text using these three operations.
Finally, we introduce how to decode the final results using schema-constraint decoding.

\subsection{Schema-Text Joint Embedding}
To capture the interaction between label schema and text, USM first learns the joint contextualized embeddings of schema labels and text tokens.
Concretely, USM first verbalizes the extraction schema $s$ as token sequence $l = \{l_{1}, l_{2}, ..., l_{|l|}\}$ following the structural schema instructor \citep{lu-etal-2022-unified}, then concatenates schema sequence $l$ and text tokens $t = \{t_{1}, t_{2}, ..., t_{|t|}\}$ as input, and finally computes the joint label-text embeddings $\mathbf{H}=[\mathbf{h}_{1}, \mathbf{h}_{2},..., \mathbf{h}_{|l|+|t|}]$ as follow:
\begin{equation}
    \mathbf{H} = \text{Encoder}(l_{1}, l_{2}, ..., l_{|l|}, t_{1}, t_{2}, ..., t_{|t|}, \mathbf{M})
\end{equation}

where $\text{Encoder}(\cdot)$ is a transformer encoder, and $\mathbf{M} \in \mathbb{R}^{(|l| + |t|) \times (|l| + |t|)}$ is the mask matrix that determines whether a pair of tokens can be attended to each other.

\subsection{Token-Token Linking for Structuring}
 
After obtaining the joint label-text embeddings $\mathbf{H}=[\mathbf{h}_{1}^{l}, ..., \mathbf{h}_{|l|}^{l}, \mathbf{h}_{1}^{t}, ..., \mathbf{h}_{|t|}^{t}]$, USM structures all valid substructures using Token-Token Linking (TTL) operations:
\begin{enumerate}[noitemsep,nolistsep]
    \item 
    \textbf{Utterance}: a continuous token sequence in the input text, e.g., entity mention ``Monet'' or event trigger ``born in''.
    We extract a single utterance with inner span head-to-tail (H2T) linking, as shown in \figurename~\ref{figure:unified-token-matching}.
    For example, to extract the span ``Monet'' and ``born in'' as valid substructures, USM utilizes H2T to link ``Monet'' to itself and link ``born'' to ``in''.
    \item
    \textbf{Association pair}: a basic related pair unit extracted from the text, e.g., relation subject-object pair (``Monet'', ``Paris'') or event trigger-argument (``born in'', ``Paris'').
    We extract span pairs with head-to-head (H2H) and tail-to-tail (T2T) linking operations.
    For example, to extract the subject-object pair ``Monet'' and ``Paris'' as a valid substructure, USM links ``Monet'' and ``Paris'' using H2H as well as links ``Monet'' and ``Paris'' using T2T.
\end{enumerate}
For the above three token-to-token linking (H2T, H2H, T2T) operations, USM respectively calculates the token-to-token linking score $\mathbf{s}_{\text{TTL}}(t_{i}, t_{j})$ over all valid token pair candidates $\langle t_{i}, t_{j} \rangle$.
For each token pair $\langle t_{i}, t_{j} \rangle$, the linking score $\mathbf{s}_{\text{TTL}}(t_{i}, t_{j})$ is calculated as:
\begin{equation}
    \begin{gathered}
    \mathbf{s}_{\text{TTL}}(t_{i}, t_{j}) = \text{FFNN}^{l}_{\text{TTL}}(\mathbf{h}_{t}^{i})^{T}\mathbf{R}_{j-i}\text{FFNN}^{r}_{\text{TTL}}(\mathbf{h}_{t}^{j})
    \end{gathered}
\end{equation}
where $\text{FFNN}^{l/r}$ are feed-forward layers with output size $d$.
$\mathbf{R}_{j-i}\in \mathbb{R}^{d \times d}$ is the rotary position embedding \citep{su-etal:reformer:2021,su-etal:global-pointer:2022} that can effectively inject relative position information into the valid structure mentioned above.

\subsection{Label-Token Linking for Utterance Conceptualizing}

Given label token embeddings $\mathbf{h}_{1}^{l}, ..., \mathbf{h}_{|l|}^{l}$ and text token embeddings $\mathbf{h}_{1}^{t}, ..., \mathbf{h}_{|t|}^{t}$, USM conceptualizes valid utterance structures with label-token linking (LTL) operations.
The output of LTL is a pair of label name and text mention, e,g., (\textit{person}, ``Monet''), (\textit{country}, ``France''), and (\textit{born}, ``born in'').
There are two types of utterance conceptualizing:
the first one is the type of mention, which indicates assigning the label types to every single mention, such as entity type \textit{person} for entity mention ``Monet'';
the second one is the predicate of object, which assigns the predicate type to each object candidate, such as relation type \textit{birth place} for ``Paris'' and event argument type \textit{place} for ``Paris''.

We conceptualize the type of mention and the predicate of object with the same label-to-token linking operation, thus enabling the two label semantics to reinforce each other.
Following the head-tail span extraction style, we name each substructure with label-to-head (L2H) and label-to-tail (L2T) linking operations.
For the pair of label name \textit{birth place} and text span \textit{Paris}, USM links the head of the label \textit{birth} with the head of text span ``Paris'' and links the tail of label \textit{place} with the tail of text span ``Paris''.

For the above two label-to-token linking (L2H, L2T) operations, USM respectively calculates the label-to-token linking score $\mathbf{s}_{\text{LTL}}(l_{i}, t_{j})$ over all valid label and text token pair candidates $\langle l_{i}, t_{j} \rangle$:
\begin{equation}
    \setlength{\abovedisplayskip}{1pt}
    \setlength{\belowdisplayskip}{1pt}
    \begin{gathered}
    \mathbf{s}_{\text{LTL}}(l_{i}, t_{j}) = 
    \text{FFNN}^{\text{label}}_{\text{LTL}}(\mathbf{h}^{l}_{i})^{T}
    \mathbf{R}_{j-i}
    \text{FFNN}_{\text{LTL}}^{\text{text}}(\mathbf{h}_{j}^{t})
    \end{gathered}
\end{equation}

\subsection{Token-Label Linking for Pairing Conceptualizing}

To conceptualize the association pair, USM links the subject of the association pair to the label name using Token-Label Linking (TLL).
Precisely, TLL operation links the subject of triplet and the predicate type with head-to-label (H2L) and tail-to-label (T2L) operations.
For instance, TLL links the head of text span ``Monet'' and the head of the label \textit{birth} with H2L and links the tail of text span ``Monet'' and the tail of the label \textit{place} with T2L following the head-tail span extraction style.
For the above two token-label linking (H2L, T2L) operations, the linking score $\mathbf{s}_{\text{TLL}}(t_{i}, l_{j})$ is computed as:
\begin{equation}
    \setlength{\abovedisplayskip}{1pt}
    \setlength{\belowdisplayskip}{1pt}
    \begin{gathered}
    \mathbf{s}_{\text{TLL}}(t_{i}, l_{j}) = \text{FFNN}^{\text{text}}_{\text{TLL}}(\mathbf{h}^{l}_{i})^{T}
    \mathbf{R}_{j-i}
    \text{FFNN}_{\text{TLL}}^{\text{label}}(\mathbf{h}_{j}^{t})
    \end{gathered}
\end{equation}

\subsection{Schema-constraint Decoding for Structure Composing}

USM decodes the final structures using a schema-constraint decoding algorithm, given substructures extracted by unified token linking operations.
During the decoding stage, we separate types for different tasks according to the schema definition.
For instance, in the joint entity and relation extraction task, we uniformly encode entity types and relation types as labels to utilize the common structuring and conceptualizing ability but compose the final result by separating the entity or relation types from input types.

As shown in \figurename~\ref{figure:unified-token-matching}, USM
1) first decodes mentions and subject-object unit extracted by token-token linking operation: \{``Monet'', ``Paris'', ``France'', (``Monet'', ``Pairs''), (``France'', ``Pairs'')\};
2) and then decodes label-mention pairs by label-token linking operation: \{(\textit{person}, ``Monet''), (\textit{country}, ``France''), (\textit{birth place}, ``Paris''), (\textit{capital}, ``Paris'')\};
3) and finally decodes label-association pairs using token-label linking operation: (``Monet'', \textit{birth place}), (``France'', \textit{capital}).
The above three token linking operations do not affect each other; hence the extraction operations are fully non-autoregressive and highly parallel.

Finally, we separate the entity types \textit{country} and \textit{person}, relation types \textit{birth place}, and \textit{capital} from input types according to the schema definition.
Based on the result from token-label linking (``Monet'', \textit{birth place}), (``France'', \textit{capital}), we can consistently obtain the full structure (``Monet'', \textit{birth place}, ``Paris'') and (``France'', \textit{capital}, ``Paris'').

\section{Learning from Heterogeneous Supervision} \label{sec:learning}

This section introduces how to leverage heterogeneous supervised resources to learn the common structuring and conceptualizing abilities for unified token linking.
Specifically, with the help of verbalized label representation and unified token linking, we unify heterogeneous supervision signals into $<$text, token pairs$>$ for pre-training.
We first pre-train the USM on the heterogeneous resources, which contain three different supervised signals, including task annotation signals (e.g., IE datasets), distant signals (e.g., distant supervision datasets), and indirect signals (e.g., question answering datasets), then adopt the pre-trained USM model to specific downstream information extraction tasks.

\subsection{Pre-training}

USM uniformly encodes label schema and text in the shared semantic representation and employs unified token linking to structure and conceptualize information from text.
To help USM to learn the common structuring and conceptualizing abilities, we collect three different supervised signals from existing linguistic sources for the pre-training of USM:

$\mathcal{D}_\text{task}$ is the task annotation dataset, where each instance has a gold annotation for information extraction.
We use Ontonotes \citep{pradhan-etal-2013-towards}, widely used in the field of information extraction as gold annotation, which contains 18 entity types.
$\mathcal{D}_\text{task}$ is used as in-task supervision signals to learn task-specific structuring and conceptualizing abilities.

  \begin{table*}[ht]
    \centering
    
    \setlength{\belowcaptionskip}{-0.3cm}

    \resizebox{.95\textwidth}{!}{

  \begin{tabular}{ccc|cc|cc|c}
  \toprule
  \textbf{Dataset} & \textbf{Metric} & \textbf{UIE} & \multicolumn{2}{c|}{\textbf{Task-specific SOTA Methods}} & \boldmath{}\textbf{USM$_\text{Roberta}$}\unboldmath{} & \textbf{USM} & \boldmath{}\textbf{USM$_\text{Unify}$}\unboldmath{} \\
  \midrule
  ACE04 & Entity F1 & 86.89  & \citep{lou-etal-2022-nested} & \textbf{87.90} & 87.79  & 87.62  & 87.34  \\
  ACE05-Ent & Entity F1 & 85.78  & \citep{lou-etal-2022-nested} & 86.91  & 86.98  & \textbf{87.14} & - \\
  CoNLL03 & Entity F1 & 92.99  & \citep{wang-etal-2021-improving} & \textbf{93.21}  & 92.76  & \textbf{93.16} & 92.97  \\
  ACE05-Rel & Relation Strict F1 & 66.06  & \citep{yan-etal-2021-partition} & 66.80  & 66.54  & \textbf{67.88} & - \\
  CoNLL04 & Relation Strict F1 & 75.00  & \citep{huguet-cabot-navigli-2021-rebel-relation} & 75.40  & 75.86  & \textbf{78.84} & 77.12  \\
  NYT   & Relation Boundary F1 & 93.54  & \citep{huguet-cabot-navigli-2021-rebel-relation} & 93.40  & 93.96  & \textbf{94.07} & 94.01  \\
  SciERC & Relation Strict F1 & 36.53  & \citep{yan-etal-2021-partition} & \textbf{38.40} & 37.05  & 37.36  & 37.42  \\
  ACE05-Evt & Event Trigger F1 & 73.36  & \citep{wang-etal-2022-query} & \textbf{73.60} & 71.68  & 72.41  & 72.31  \\
  ACE05-Evt & Event Argument F1 & 54.79  & \citep{wang-etal-2022-query} & 55.10  & 55.37  & \textbf{55.83} & 53.57  \\
  CASIE & Event Trigger F1 & 69.33  & \citep{lu-etal-2021-text2event} & 68.98  & 70.77  & \textbf{71.73} & 71.56  \\
  CASIE & Event Argument F1 & 61.30  & \citep{lu-etal-2021-text2event} & 60.37  & 63.05  & \textbf{63.26} & 63.00  \\
  14-res & Sentiment Triplet F1 & 74.52  & \citep{lu-etal-2022-unified}     & 74.52  & 76.35  & \textbf{77.26} & 77.29  \\
  14-lap & Sentiment Triplet F1 & 63.88  & \citep{lu-etal-2022-unified}     & 63.88  & 65.46  & \textbf{65.51} & 66.60  \\
  15-res & Sentiment Triplet F1 & 67.15  & \citep{lu-etal-2022-unified}     & 67.15  & 68.80  & \textbf{69.86} & - \\
  16-res & Sentiment Triplet F1 & 75.07  & \citep{lu-etal-2022-unified}     & 75.07  & 76.73  & \textbf{78.25} & - \\
  \midrule
  AVE-unify & -     & 71.10  & -     & 71.34  & 71.83  & \textbf{72.46} & 72.11  \\
  AVE-total & -     & 71.75  & -     & 72.05  & 72.61  & \textbf{73.35} & - \\
  \bottomrule
  \end{tabular}%

  }
    \caption{
    Overall results of USM on different datasets.
    AVE-unify indicates the average performance of non-overlapped datasets (except ACE05-Rel/Evt and 15/16-res), and AVE-total indicates the average performance of all datasets.}
   
    \label{tab:overall}%
  \end{table*}%

$\mathcal{D}_\text{distant}$ is the distant supervision dataset, where each instance is aligned by text and knowledge base.
Distant supervision is a common practice to obtain large-scale training data for information extraction \citep{mintz-etal-2009-distant,riedel-etal-2013-relation}.
We employ NYT \citep{riedel-etal-2013-relation} and Rebel \citep{huguet-cabot-navigli-2021-rebel-relation} as our distant supervision datasets, which are obtained by aligning text with Freebase and Wikidata, respectively.
Rebel dataset has a large label schema, and all verbalized schemas are too long to be concatenated with input text and fed to the pre-trained transformer encoder.
We sample negative label schema to construct meta schema \citep{lu-etal-2022-unified} as label schema for pre-training.

$\mathcal{D}_\text{indirect}$ is the indirect supervision dataset, where each instance is derived from other related NLP tasks \citep{NEURIPS2020_67ff32d4,chen-etal-2022-new}.
We utilize reading comprehension datasets from MRQA \citep{fisch-etal-2019-mrqa} as our indirect supervision datasets: HotpotQA \citep{yang-etal-2018-hotpotqa}, Natural Questions \citep{kwiatkowski-etal-2019-natural}, NewsQA \citep{trischler-etal-2017-newsqa}, SQuAD \citep{rajpurkar-etal-2016-squad} and TriviaQA \citep{joshi-etal-2017-triviaqa}.
Compared with limited entity types in $\mathcal{D}_\text{task}$  and relation types $\mathcal{D}_\text{distant}$, diversified question expressions can provide richer label semantic information for learning conceptualizing.
For each (question, context, answer) instance in $\mathcal{D}_\text{indirect}$, we take the question as label schema, the context as input text, and the answer as mention.
It captures structuring and conceptualizing ability in the pre-training stage by learning token-token and label-token linking operations.

\subsection{Learning function}
For pre-training, fine-tuning and multi-task learning, we unify all datasets as $\{(x_{i}, y_{i})\}$, where $x_{i}$ is text and $y_{i}$ is linking annotation of each token linking pair (TTM, LTM, TLM).
We use the same learning function for all settings with the homogenized data format.

The main challenge of USM learning is the sparsity of linked token pairs.
The linked ratio only occupies less than 1\% of all valid token pair candidates.
To overcome the extreme sparsity of linking instances, we optimize class imbalance loss \cite{su-etal:global-pointer:2022} for each instance as follows:
\begin{equation}
\begin{aligned}
    \mathcal{L} = \sum_{m \in \mathcal{M}} & \log \left(1 + \sum_{(i,j) \in m^{+}} e^{-s_{m}(i, j)} \right) \\
    & + \log \left(1 + \sum_{(i,j) \in m^{-}} e^{s_{m}(i, j)}\right)
\end{aligned}
\end{equation}
where $\mathcal{M}$ denotes linking types of USM, $m^{+}$ indicates the linked pairs, $m^{-}$ indicates the non-linked pairs, and $s_{m}(i, j)$ is the predicate linking score for the linking operation $m$.

\section{Experiments} \label{sec:experiments}

This section conducts massive experiments under supervised settings and transfer settings to demonstrate the effectiveness of the proposed unified semantic matching framework.

  \begin{table*}[ht]
    \centering

    \setlength{\belowcaptionskip}{-0.3cm}

    \resizebox{.95\textwidth}{!}{

  \begin{tabular}{c|ccccccccc}
  \toprule
        & \textbf{Movie} & \textbf{Restaurant} & \textbf{Social} & \textbf{AI} & \textbf{Literature} & \textbf{Music} & \textbf{Politics} & \textbf{Science} & \textbf{Ave} \\
  \midrule
  \multicolumn{1}{l}{} & \multicolumn{9}{c}{\boldmath{}\textbf{Performance on Unseen Label Subset of $\mathcal{D}_{t}$ and $\mathcal{D}_{i}$}\unboldmath{}} \\
  \midrule
  \textbf{\#Unseen/\#All} & 12/12 & 7/8   & 7/10  & 10/14 & 8/12  & 9/13  & 5/9   & 13/17 & - \\
  \midrule
  \multicolumn{1}{l|}{$\mathcal{D}_{\text{task}}$} & 25.07 & 2.50   & 22.54 & 10.82 & 50.74 & 44.11 & 9.75  & 13.98 & 22.44  \\
  \multicolumn{1}{l|}{$\mathcal{D}_{\text{task}}$ + $\mathcal{D}_{\text{indirect}}$} & 37.73  & 14.73  & 29.34  & 28.18  & 56.00  & 44.93  & 36.10  & 44.09  & 36.39  \\
  \midrule
  \multicolumn{1}{l}{} & \multicolumn{9}{c}{\textbf{Performance on Unseen Label Subset of Pre-training Dataset}} \\
  \midrule
  \textbf{\#Unseen/\#All} & 10/12 & 7/8   & 6/10  & 8/14  & 7/12  & 8/13  & 4/9   & 12/17 & - \\
  \midrule
  \multicolumn{1}{l|}{$\mathcal{D}_{\text{task}}$} & 32.1  & 2.50   & 1.64  & 10.68 & 52.42 & 45.93 & 11.16 & 14.12 & 21.32  \\
  \multicolumn{1}{l|}{$\mathcal{D}_{\text{task}}$ + $\mathcal{D}_{\text{indirect}}$} & 39.76  & 14.73  & 20.62  & 24.12  & 56.24  & 44.21  & 32.92  & 44.25  & 34.61  \\
  \multicolumn{1}{l|}{$\mathcal{D}_{\text{task}}$ + $\mathcal{D}_{\text{distant}}$} & 35.35  & 21.10  & 40.64  & 27.57  & 56.97  & 49.29  & 43.72  & 44.05  & 39.84  \\
  \multicolumn{1}{l|}{$\mathcal{D}_{\text{task}}$ + $\mathcal{D}_{\text{distant}}$ +  $\mathcal{D}_{\text{indirect}}$} & 42.11  & 26.01  & 44.37  & 34.91  & 65.69  & 60.07  & 56.65  & 55.26  & 48.13  \\
  \midrule
  $\mathbf{\Delta}$ & 10.01 & 23.51 & 42.73 & 24.23 & 13.27 & 14.14 & 45.49 & 41.14 & 26.82  \\
  \bottomrule
  \end{tabular}%

}
    \caption{
        Performance of Zero-shot transfer settings on unseen entity label subset with different supervision signals.
        Unseen indicates label types that do not appear in the pre-training dataset.
        $\mathbf{\Delta}$ indicates the improvement of pre-training using extra supervision signals ($\mathcal{D}_{\text{distant}}$ and $\mathcal{D}_{\text{indirect}}$).
   }
    \label{tab:signals}%
  \end{table*}%

  \begin{table}[htbp]
    \centering
    
    \setlength{\belowcaptionskip}{-0.3cm}

    \resizebox{.35\textwidth}{!}{
      \begin{tabular}{lcc}
      \toprule
            & CoNLL04 & Model Size \\
      \midrule
      GPT-3 & 18.10  & 137B \\
      \textsc{DeepStruct} & 25.80  & 10B \\
      \midrule
      USM & \textbf{25.95}  & 356M \\
      \bottomrule
      \end{tabular}
      }
    \caption{
        Performance of Zero-shot transfer settings on relation extraction.
        * GPT-3 175B indicates formulating the extraction task as a question answering problem through prompting, and \textsc{DeepStruct} 10B is a pre-trained language model for structure prediction \citep{wang-etal-2022-deepstruct}
    }
    \label{tab:zero-shot-relation}
  \end{table}%

  \begin{table}[ht]
    \centering

    \setlength{\belowcaptionskip}{-0.5cm}

\resizebox{0.48\textwidth}{!}{

  \begin{tabular}{cccccc}
  \toprule
        & \textbf{Model} & \textbf{1-Shot} & \textbf{5-Shot} & \textbf{10-Shot} & \textbf{AVE-S} \\
  \midrule
  \multicolumn{1}{c}{\multirow{4}[2]{*}{\shortstack{Entity\\ CoNLL03}}} & UIE-Large* & 57.53  & 75.32  & 79.12  & 70.66  \\
        & USM$_\text{Roberta}$ & 9.69  & 40.66  & 62.87  & 37.74  \\
        & USM$_\text{Symbolic}$ & 60.56  & 81.87  & 83.87  & 75.43  \\
        & USM   & \textbf{71.11} & \textbf{83.25} & \textbf{84.58} & \textbf{79.65} \\
  \midrule
  \multicolumn{1}{c}{\multirow{4}[2]{*}{\shortstack{Relation\\CoNLL04}}} & UIE-Large* & 34.88  & 51.64  & 58.98  & 48.50  \\
        & USM$_\text{Roberta}$ & 0.00  & 12.81  & 31.02  & 14.61  \\
        & USM$_\text{Symbolic}$ & 13.45  & 48.31  & 58.91  & 40.22  \\
        & USM   & \textbf{36.17} & \textbf{53.20} & \textbf{60.99} & \textbf{50.12} \\
  \midrule
  \multicolumn{1}{c}{\multirow{4}[2]{*}{\shortstack{Event \\ Trigger \\ ACE05-Evt}}} & UIE-Large* & \textbf{42.37}  & 53.07  & 54.35  & 49.93  \\
        & USM$_\text{Roberta}$ & 26.39  & 47.10  & 51.46  & 41.65  \\
        & USM$_\text{Symbolic}$ & 1.97  & 30.77  & 52.30  & 28.35  \\
        & USM   & 40.86 & \textbf{55.61} & \textbf{58.79} & \textbf{51.75} \\
  \midrule
  \multicolumn{1}{c}{\multirow{4}[2]{*}{\shortstack{Event \\ Argument \\ ACE05-Evt}}} & UIE-Large* & 14.56  & 31.20  & 35.19  & 26.98  \\
        & USM$_\text{Roberta}$ & 6.47  & 27.00  & 34.20  & 22.56  \\
        & USM$_\text{Symbolic}$ & 0.08  & 13.71  & 33.52  & 15.77  \\
        & USM   & \textbf{19.01} & \textbf{36.69} & \textbf{42.48} & \textbf{32.73} \\
  \midrule
  \multicolumn{1}{c}{\multirow{4}[2]{*}{\shortstack{Sentiment \\ 16res}}} & UIE-Large* & 23.04  & 42.67  & 53.28  & 39.66  \\
        & USM$_\text{Roberta}$ & 2.68  & 35.71  & 48.56  & 28.98  \\
        & USM$_\text{Symbolic}$ & 20.08  & 41.25  & 50.90  & 37.41  \\
        & USM   & \textbf{30.81} & \textbf{52.06} & \textbf{58.29} & \textbf{47.05} \\
  \bottomrule
  \end{tabular}%

}

    \caption{Few-shot results on end-to-end IE tasks.
    For a fair comparison, we conduct text-structure pre-training from T5-v1.1-large using the same pre-training corpus of USM, refer to UIE-Large*.
    }
    \label{tab:few-shot}%
  \end{table}%

\subsection{Experiments on Supervised Settings}
We conduct supervised experiments on extensive information extraction tasks, including 4 tasks and 13 datasets (entity extraction, relation extraction, event extraction, sentiment extraction) and their combinations (e.g., joint entity-relation extraction).
The used datasets includes ACE04 \citep{ace2004-annotation}, ACE05 \citep{ace2005-annotation}; CoNLL03 \citep{tjongkimsang2003conll}, CoNLL04 \citep{roth-yih-2004-linear}, SciERC \citep{luan-etal-2018-multi}, NYT \citep{10.1007/978-3-642-15939-8_10}, CASIE \citep{Satyapanich_Ferraro_Finin_2020}, SemEval-14/15/16 \citep{pontiki-etal-2014-semeval,pontiki-etal-2015-semeval,pontiki-etal-2016-semeval}.
We employ the same end-to-end settings and evaluation metrics as  \citet{lu-etal-2022-unified}.

We compare the proposed USM framework with the task-specific state-of-the-art methods and the unified structure generation method -- UIE \citep{lu-etal-2022-unified}.
For our approach, we show three different settings:
\begin{itemize}[noitemsep,nolistsep]
    \item USM is the pre-trained model which learned unified token linking ability from heterogeneous supervision;
    \item USM$_{\text{Roberta}}$ is the initial model of the pre-trained USM, which employs RoBERTa-Large \citep{roberta} as the pre-trained transformer encoder;
    \item USM$_{\text{Unify}}$ is initialized by the pre-trained USM and conducts multi-task learning with all datasets but ignores overlapped datasets: ACE05-Ent/Rel and 15/16-res.
\end{itemize}
For the USM$_{\text{Roberta}}$ and USM settings, we fine-tune them on each specific task separately.
We run each experiment with three seeds and report their average performance.

\tablename~\ref{tab:overall} shows the overall performance of USM and other baselines on the 13 datasets, where AVE-unify indicates the average performance of non-overlapped datasets, and AVE-total indicates the average performance of all datasets. We can observe that:
    1) \textit{By verbalizing labels and modeling all IE tasks as unified token linking, USM provides a novel and effective framework for IE.}
    USM achieves state-of-the-art performance and outperforms the strong task-specific methods by 1.30 in AVE-total.
    Even without pre-training, USM$_{\text{Roberta}}$ also shows strong performance, which indicates the strong portability and generalization ability of unified token linking.
    2) \textit{Heterogeneous supervision provides a better foundation for structuring and conceptualizing information extraction.}
    Compared to the initial model USM$_{\text{Roberta}}$ and the pre-trained model USM, the heterogeneous pre-training achieved an average 0.74 improvement across all datasets.
    3) \textit{By homogenizing diversified label schemas and heterogeneous target structures into the unified token sequence, USM$_\text{Unify}$ can solve massive IE tasks with a single multi-task model.}
    USM$_\text{Unify}$ outperforms task-specific state-of-the-art methods with different model architectures and encoder backbones in average, providing an efficient solution for application and deployment.

\subsection{Experiments on Zero-shot Transfer Settings}
We conduct zero-shot cross-type transfer experiments on 9 datasets across various domains to verify the transferable conceptualization learned by USM.
In this setting, we directly employ the pre-trained USM to conduct extraction on new datasets.

For entity extraction, the cross-type extraction datasets include Movie (MIT-Movie), Restaurant (MIT-Restaurant) \citep{6639301}, Social (WNUT-16) \citep{strauss-etal-2016-results}, and AI/Literature/Music/Politics/Science from CrossNER \citep{Liu_Xu_Yu_Dai_Ji_Cahyawijaya_Madotto_Fung_2021}.
We investigate the effect of different supervised signals in the zero-shot entity extraction setting.
$\mathcal{D}_{task}$ indicates we first train USM on the common entity extraction dataset -- Ontonotes, then directly conduct extraction on the new types, which emulates the most common label transfer method used in real-world scenarios.
To be consistent with the real scenario, we select the best checkpoint according to the F1 score on the dev set of $\mathcal{D}_{task}$.

For zero-shot relation extraction, we compare USM with the following strong baselines:
\begin{itemize}[noitemsep,nolistsep]
    \item GPT-3 175B \citep{NEURIPS2020_1457c0d6} is a large-scale, generative pre-trained model, which can extract entity and relation by formulating the task as a question answering problem through prompting \citep{wang-etal-2022-deepstruct}.
    \item \textsc{DeepStruct} 10B is a structured prediction model pre-trained on six large-scale entity, relation, and triple datasets \citep{wang-etal-2022-deepstruct}.
\end{itemize}

\tablename~\ref{tab:signals} shows the entity extraction performance on the unseen label subset, in which types are not appearing in the pre-training dataset.
And \tablename~\ref{tab:zero-shot-relation} shows the performance of zero-shot relation extraction on CoNLL04.
From \tablename~\ref{tab:signals} and \tablename~\ref{tab:zero-shot-relation}, we can see that:
    1) \textit{USM has a strong zero-shot transferability across labels.}
    USM shows good migration performance on Movie, Literature, and Music domains even when learning from $\mathcal{D}_{\text{task}}$ with limited entity types.
    For relation extraction, USM (356M) outperforms the strong zero-shot baseline GPT-3 (175B) and  \textsc{DeepStructure} (10B) with a smaller model size.
    2) \textit{Heterogeneous supervision boosts USM with unified label semantics and outperforms the task annotation baseline by a large margin.}
    Compared to the task annotation baseline ($\mathcal{D}_{\text{task}}$), USM significantly and consistently improves the performance on all datasets.

\subsection{Experiments on Few-shot Transfer Settings}

To further investigate the effects of verbalized label semantics, we conduct few-shot transfer experiments on four IE tasks and compare USM with the following baselines:
\begin{itemize}[noitemsep,nolistsep]
    \item \textbf{UIE-large*} is the pre-trained sequence-to-structure model for effective low-resource IE tasks, which injects label semantics by generating labels and words in structured extraction language synchronously and guiding the generation with a structural schema instructor.
    \item \textbf{USM$_\text{Roberta}$} is the initial model of USM, which directly use Roberta-large as the pre-trained encoder;
    \item \textbf{USM$_\text{Symbolic}$} replaces the names of labels with symbolic representation (meaning-less labels, e.g., label1, label2, ...) during the fine-tuning stage of USM, which is used to verify the effect of verbalized label semantics.
\end{itemize}

For few-shot transfer experiments, we follow the data splits and settings with the previous work \citep{lu-etal-2022-unified} and repeat each experiment 10 times to avoid the influence of random sampling \citep{huang-etal-2021-shot}.
\tablename~\ref{tab:few-shot} shows the performance on 4 IE tasks under the few-shot settings, where AVE-S is the average performance of 1/5/10-shot experiments.
We can see that:
    1) \textit{By modeling IE tasks via unified semantic matching, USM exceeds the few-shot state-of-the-art UIE-large 5.11 on average.}
    Although UIE also adopts verbalized label representation, this structure generation method needs to learn to generate the novel schema word in the target structure during transfer learning.
    In contrast, USM only needs to learn to match them, providing a better inductive bias and leading to a much smaller decoding search space.
    The pre-trained unified token linking ability boosts the USM in all settings.
    2) \textit{It is crucial to verbalize label schemas rather than meaningless symbols, especially for complex extraction tasks.}
    USM$_{\text{Symbolic}}$, which uses symbolic labels instead of verbalized labels, drastically reduces performance on all tasks.
    For tasks with more semantic types, such as event extraction with 33 types, the performance drops significantly, even lower than that of USM$_{\text{Roberta}}$ initialized directly with Roberta-large.

\section{Related Work}

In the past decade, due to powerful representation ability, deep learning methods \citep{10.5555/944919.944966,10.5555/1953048.2078186} have made amazing achievements in information extraction tasks.
Most of these methods decompose extraction into multiple sub-tasks and follow the classical neural classifier method \citep{NIPS2012_c399862d} to model each sub-task, such as entity extraction, relation classification, event trigger detection, event argument classification, etc.
And several architectures are proposed to model the extraction, such as sequence tagging \citep{lample-etal-2016-neural,zheng-etal-2017-joint}, span classification \citep{sohrab-miwa-2018-deep,song-etal-2019-leveraging,wadden-etal-2019-entity}, table filling \citep{gupta-etal-2016-table,wang-lu-2020-two}, question answering \citep{levy-etal-2017-zero,li-etal-2020-unified}, and token pair \citep{wang-etal-2020-tplinker,yu-etal-2021-maximal}.

Recently, to solve various IE tasks with a single architecture, UIE employs unified structure generation, models the various IE tasks with structured extraction language, and pre-trains the ability of structure generation using distant text-structure supervision \citep{lu-etal-2022-unified}.
Unlike the generation-based approach, we model universal information extraction as unified token linking, which reduces the search space during decoding and leads to better generalization performance.
Beyond distant supervision, we further introduce indirect supervision from related NLP tasks to learn the unified token linking ability.

Similar to USM in this paper, matching-based IE approaches aim to verbalize the label schema and structure candidate to achieve better generalization \citep{liu-etal-2022-pre}.
Such methods usually use pre-extracted syntactic structures \citep{wang-etal-2021-zero} and semantic structures \citep{huang-etal-2018-zero} as candidate structures, then model the extraction as text entailment \citep{obamuyide-vlachos-2018-zero,sainz-etal-2021-label,lyu-etal-2021-zero,sainz-etal-2022-textual} and semantic structure mapping \citep{chen-li-2021-zs,dong-etal-2021-mapre}.
Different from the pre-extraction and matching style, this paper decouples various IE tasks to unified token linking operations and designs a one-pass end-to-end information extraction framework for modeling all tasks.

\section{Conclusion}

In this paper, we propose a unified semantic matching framework – USM, which jointly encodes extraction schema and input text, uniformly extracts substructures in parallel, and controllably decodes target structures on demand.
Experimental results show that USM achieves state-of-the-art performance under the supervised experiments and shows strong generalization ability under zero/few-shot transfer settings, which verifies USM is a novel, transferable, controllable, and efficient framework. 
For future work, we want to extend USM to NLU tasks, e.g., text classification, and investigate more indirect supervision signals for IE, e.g., text entailment.

\section{Acknowledgments}
We sincerely thank the reviewers for their insightful comments and valuable suggestions.
This work is supported by the National Key Research and Development Program of China (No.2020AAA0109400) and the Natural Science Foundation of China (No.62122077, 61876223, and 62106251).
Hongyu Lin is sponsored by CCF-Baidu Open Fund.

\appendix

\section{Appendix: Experiment Details}
This section describes the details of the experiments, including implementation details and extra experiments analysis.

\subsection{Implementation Details}
For all experiments, we optimize our model using AdamW \citep{loshchilov2018decoupled} with the constant learning rate.
For single-task fine-tuning, we tune the learning rate from \{1e-5, 2e-5, 3e-5\} with three seeds and select the best hyper-parameter setting according to the performance of the development set.
For multi-task learning of USM$_\text{Unify}$, we select the best checkpoint according to the average performance of all datasets.
We conducted each experiment on NVIDIA A100 GPUs, and 
detailed hyper-parameters are shown in \tablename~\ref{tab:hyper-para}.

  \begin{table}[ht]
    \centering
    \resizebox{.45\textwidth}{!}{

      \begin{tabular}{lccc}
      \toprule
            & \textbf{Learning Rate} & \textbf{Global Batch} & \textbf{Epoch} \\
      \midrule
      \textit{\textbf{Pre-training}} & 2e-5  & 96    & 5 \\
      \midrule
      \textit{\textbf{Fine-tuning}} &       &       &  \\
      ~ Entity & 1e-5, 2e-5, 3e-5 & 64    & 100 \\
      ~ Relation & 1e-5, 2e-5, 3e-5 & 64    & 200 \\
      ~ Event & 1e-5, 2e-5, 3e-5 & 96    & 200 \\
      ~ Sentiment & 1e-5, 2e-5, 3e-5 & 32    & 100 \\
      ~ Low-resource & 2e-5 & 32 & 200 \\
      \midrule
      \textit{\textbf{Multi-task}} & 2e-5  & 96    & 200 \\
      \bottomrule
      \end{tabular}%
    
    }
    \caption{Hyper-parameters of USM experiments.}
    \label{tab:hyper-para}%
  \end{table}%

\subsection{Pre-train Datasets}

We collect three types of supervision signals for model pre-training: named entity annotation in Ontonotes for task annotation $\mathcal{D}_{\text{task}}$; NYT \citep{10.1007/978-3-642-15939-8_10} and Rebel \citep{huguet-cabot-navigli-2021-rebel-relation} for distant supervision $\mathcal{D}_{\text{distant}}$; machine reading comprehension from MRQA \citep{fisch-etal-2019-mrqa} for indirect supervision $\mathcal{D}_{\text{indirect}}$.
For the Rebel data, we only keep the 230 most frequently occurring relation types and randomly sample 300K instances for pre-training.
For the reading comprehension data, we reserve a maximum of 5 questions for each instance and filter out instances where the total tokenized length of question and context exceeds 500.
The final statistics are shown in \tablename~\ref{tab:pretrain-datasets}.

  \begin{table}[htbp]
    \centering
      \begin{tabular}{lll}
      \toprule
            & \textbf{Dataset} & \textbf{\#instance} \\
      \midrule
      $\mathcal{D}_{\text{task}}$ & Ontonote & 60K \\
      $\mathcal{D}_{\text{distant}}$ & NYT + Rebel & 356K \\
      $\mathcal{D}_{\text{indirect}}$ & MRQA  & 195K \\
      \bottomrule
      \end{tabular}%
    \caption{Detailed statistics of pre-training datasets.}
    \label{tab:pretrain-datasets}%
  \end{table}%

\subsection{Ablation Analysis of Label-Text Interaction} \label{sec:interaction-label-text}

\begin{figure}[H]
\centering
 \subfloat[Label $\Leftrightarrow$ Text: Label and text can attend to each other.]{
	\begin{minipage}[c][1\width]{
	   0.225\textwidth}
	   \centering
	   \includegraphics[width=1\textwidth]{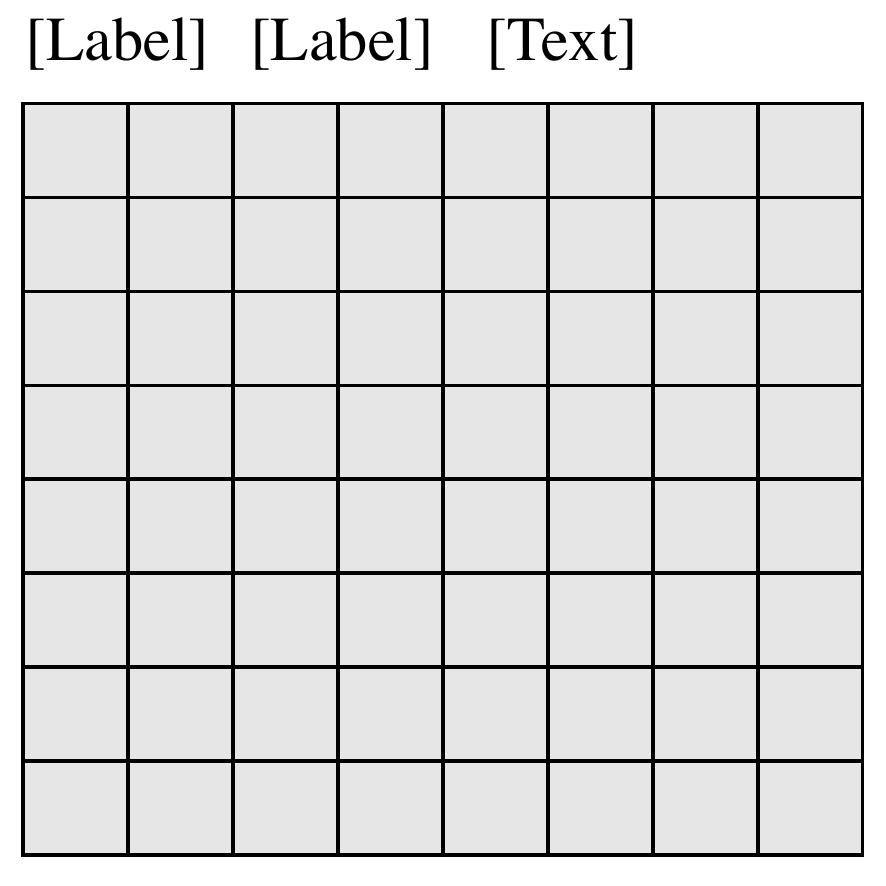}
	   \label{figure:full_attention_mask}
	\end{minipage}}
 \hfill
 \subfloat[Label $\times$ Text: Label and text can not attend to each other.]{
	\begin{minipage}[c][1\width]{
	   0.225\textwidth}
	   \centering
	   \includegraphics[width=1\textwidth]{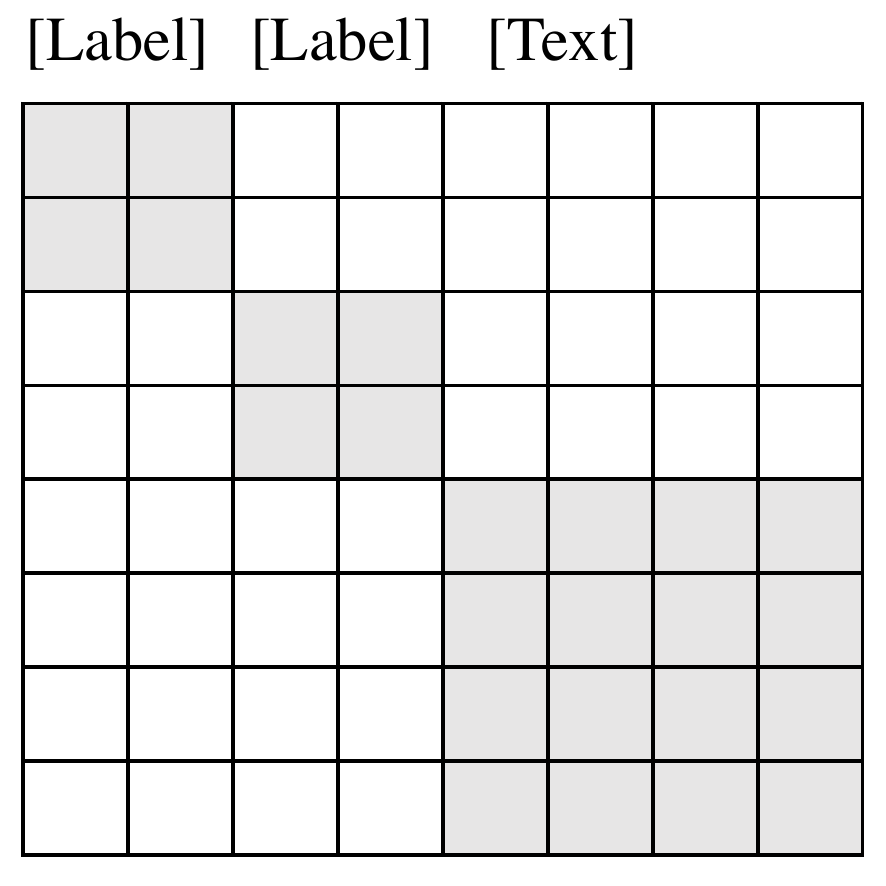}
	   \label{figure:partial_attention_mask}
	\end{minipage}}
\caption{Different attention masks for text-schema joint embedding.}
\label{figure:attention_mask}
\end{figure}

To investigate the effect of label-text interaction and accelerate the extraction process, we propose an approximate shallow label-text interaction model to reuse the computation of label embedding during the inference stage.
Motivated by \citet{unilm}, we design attention mask strategies to control the interaction between label and text, as illustrated in \figurename~\ref{figure:attention_mask}.
In the full mask setting (Label $\Leftrightarrow$ Text, \figurename~\ref{figure:full_attention_mask}), label and text can attend to each other to obtain deep interaction;
in the partial mask setting (Label $\times$ Text, \figurename~\ref{figure:partial_attention_mask}), label and text only attend to themselves. 
For the partial mask setting, USM can cache and reuse the calculation of label embedding to reduce the computation cost in a dual encoder way during the inference stage.

  \begin{table}[ht]
    \centering
    \resizebox{.48\textwidth}{!}{

  \begin{tabular}{ccccc}
  \toprule
        & \textbf{Entity} & \textbf{Relation} & \textbf{Event} & \multicolumn{1}{c}{\textbf{Sentiment}} \\
  \midrule
  \multicolumn{5}{c}{\textbf{Full-shot}} \\
  \midrule
  Label $\Leftrightarrow$ Text & 97.03  & 81.91  & 63.51  & 81.22 \\
  Label $\times$ Text & 96.99  & 81.18  & 62.03  & 80.92 \\
  \midrule
  \multicolumn{5}{c}{\textbf{Few-shot (AVE-S)}} \\
  \midrule
  Label $\Leftrightarrow$ Text & 82.12  & 52.23  & 37.52  & 51.51 \\
  Label $\times$ Text & 82.37  & 45.75  & 24.70  & 26.65 \\
  \bottomrule
  \end{tabular}%

      }
    \caption{Experiment results on the development set of entity (CoNLL03), relation (CoNLL04), event (ACE05-Evt argument) and sentiment (16res) of USM with different label-text interaction.}
    \label{tab:interaction}%
  \end{table}%

\tablename~\ref{tab:interaction} shows the performance of two different label-text interactions, and we can see that:
1) Deep interaction ($\Leftrightarrow$) can effectively improve the ability of unified token linking, especially in low-resource settings.
2) In resource-rich scenarios, shallow interaction ($\times$) can replace deep interaction between label-text linking.
This dynamic and variable scalability enables USM to have better application scenarios in practice:
for common rich resource extraction tasks, USM can pre-compute the representation of label and text separately in a dual encoder fashion, speeding up the inference process without the need for other deployments;
for low-resource extraction tasks, USM can use deep-level interactive information to improve transfer ability and retain high parallelism.

\subsection{Effects of Controllable Ability}

To investigate the controllable ability of USM, we conduct partial extraction experiments on the CoNLL04 (Joint Entity and Relation Extraction), ACE05-Evt (Event Trigger and Argument), and 14lap (Sentiment Extraction).
We employ two kinds of partial extraction settings:
1) partial task extraction: we train an end-to-end joint entity and relation extraction model using the full schema of CoNLL04 (entity and relation) but feed the partial schema (entity) to USM.
2) partial label extraction: we train an extraction model on the full label set (\textit{positive}, \textit{neutral}, \textit{negative} of sentiment), and only extract part of the label set (\textit{positive}) from the text.
\tablename~\ref{tab:controllable} shows the performance of three different partial extraction experiments.
We can see that USM achieves almost the same performance in both settings and has highly controllable extraction ability.

\begin{table}[htbp]
  \centering
  \resizebox{0.48\textwidth}{!}{

    \begin{tabular}{lccl}
    \toprule
          & \textbf{Full}  & \textbf{Partial} & \textbf{Partial Details} \\
    \midrule
    CoNLL04 Entity & 90.74  & 90.50  & Only Entity \\
    ACE05-Evt Trigger & 70.40  & 70.99  & Only 16 Types of 33 Types \\
    ACE05-Evt Argument & 60.87  & 60.24  & Only 16 Types of 33 Types \\
    14lap Sentiment & 75.00  & 74.78  & Only Positive of 3 Types \\
    \bottomrule
    \end{tabular}%

    }
  \caption{
    Experiment results of partial extraction schema on the development set of different datasets.
    Partial indicates feeding part of the whole schema to USM, such as only extracting \textit{positive} sentiment rather than extracting all types (\textit{positive}, \textit{neutral}, \textit{negative}) from the text.
    All results are evaluated on the partial extraction schema.
    For instance, the performances of ACE05-Evt Trigger under the full and partial settings result from 16 types in the partial extraction schema.
    }
  \label{tab:controllable}%
\end{table}%

\begin{table}[htbp]
  \centering
  \resizebox{0.46\textwidth}{!}{

    \begin{tabular}{lcccc}
    \toprule
          & \multicolumn{1}{l}{\textbf{14res}} & \multicolumn{1}{l}{\textbf{14lap}} & \multicolumn{1}{l}{\textbf{15res}} & \multicolumn{1}{l}{\textbf{16res}} \\
    \midrule
    \textbf{\textit{System using BERT-base}} \\
    \citep{xu-etal-2020-position} & 62.40  & 51.04 & 57.53 & 63.83 \\
    \citep{xu-etal-2021-learning} & \textbf{71.85} & 59.38 & 63.27 & 70.26 \\
    \citep{10.1145/3459637.3482058} & 69.61 & \textbf{59.50}  & 62.72 & 68.41 \\
    \citep{chen-etal-2022-enhanced} & 71.78 & 58.81 & 61.93 & 68.33 \\
    \midrule
    USM$_\text{BERT-base}$ & \textbf{71.87} & 58.63 & \textbf{63.41} & \textbf{72.68} \\
    \bottomrule
    \end{tabular}%

    }
  \caption{Experiment results of USM$_\text{BERT-base}$ on aspect based sentiment triplet extraction tasks.}
  \label{tab:absa-base}%
\end{table}%

\begin{table}[htbp]
  \centering
\begin{tabular}{lrrr}
\toprule
      & \multicolumn{1}{l}{\textbf{P}} & \multicolumn{1}{l}{\textbf{R}} & \multicolumn{1}{l}{\textbf{F}} \\
\midrule
\textbf{\textit{System using BERT-base}} \\
\citep{wang-etal-2020-tplinker} & 91.4  & \textbf{92.6}  & 92.0 \\
\citep{sui:2020:spn} & 92.5  & 92.2  & 92.3 \\
\citep{zheng-etal-2021-prgc} & 93.5  & 91.9  & 92.7 \\
\midrule
USM$_{\text{BERT-base}}$   & \textbf{93.7 }& 91.9  & \textbf{92.8} \\
\bottomrule
\end{tabular}%

  \caption{
    Experiment results of USM$_{\text{BERT-base}}$ on the NYT.
    }
  \label{tab:nyt-base}%
\end{table}%

\subsection{Comparison of BERT-base}
This section compares USM with other BERT-base based state-of-the-art systems.
USM$_\text{BERT-base}$ indicates USM uses BERT-base \citep{devlin-etal-2019-bert} as a pre-trained transformer encoder.
\tablename~\ref{tab:absa-base} shows the performance of USM and the state-of-the-art systems on the four aspect-based sentiment analysis datasets, and \tablename~\ref{tab:nyt-base}  shows the performance of USM and the state-of-the-art joint entity relation extraction systems on the NYT dataset.
We can see that USM$_\text{BERT-base}$ achieves competitive performance on above datasets, which verifies the effectiveness of the proposed unified semantic matching framework.

\subsection{Effect of Token-Label Linking}

This section investigates the effect of the token-label linking operation.
\tablename~\ref{tab:golden-links} shows results of different decoding strategies with golden token links: 
1) \textit{Full} employs all three types of token linking operations to decode the final structures;
2) \textit{w/o} TLL indicates decoding without the token-label links for pairing conceptualizing.

\begin{table}[htbp]
  \centering
  \resizebox{0.45\textwidth}{!}{

    \begin{tabular}{cccc}
    \toprule
    \textbf{Dataset} & \textbf{Metric} & \multicolumn{2}{c}{\textbf{F1 with golden links}} \\
          &       & \textit{w/o} TLL & \textit{Full} \\
    \midrule
    ACE05-Rel & Relation Strict F1 & 98.54  & 99.96  \\
    CoNLL04 & Relation Strict F1 & 100.00  & 100.00  \\
    NYT   & Relation Boundary F1 & 72.74  & 100.00  \\
    SciERC & Relation Strict F1 & 92.06  & 99.74  \\
    ACE05-Evt & Event Argument F1 & 98.75  & 100.00  \\
    CASIE & Event Argument F1 & 99.98  & 99.99  \\
    14-res & Sentiment Triplet F1 & 99.10  & 100.00  \\
    14-lap & Sentiment Triplet F1 & 98.54  & 100.00  \\
    \bottomrule
    \end{tabular}%

  }
  \caption{
  Performance of different decoding strategies using golden links.
  }
  \label{tab:golden-links}%
\end{table}%

\bibliography{usm}

\end{document}